\documentclass{ecai2004}
\usepackage{times}
\usepackage{graphicx}
\usepackage{latexsym}

\begin{document}
\title{Local search heuristics: Fitness Cloud versus Fitness Landscape}
\author{Collard Philippe \and Verel  S\'ebastien \and Clergue Manuel 
\institute{University of Nice-Sophia Antipolis,
I3S Laboratory, France, email:{pc@unice.fr verel@i3s.unice.fr clerguem@i3s.unice.fr}}
}
\maketitle
\bibliographystyle{ecai2004}

\begin{abstract}
This paper introduces the concept of \textit{fitness cloud} as an alternative way to visualize and analyze search spaces than given by the geographic notion of fitness landscape. It is argued that the fitness cloud concept overcomes several deficiencies of the landscape representation. Our analysis is based on the correlation between fitness of solutions and fitnesses of nearest solutions according to some neighboring. We focus on the behavior of local search heuristics, such as \textit{hill climber}, on the well-known NK fitness landscape. In both cases the fitness vs. fitness correlation is shown to be related to the epistatic parameter $K$. 
\end{abstract}

%%% -------------------------------------------------------------------
\section*{Introduction}
The fitness landscape has first been introduced in 1932 by the biologist Wright (\cite{wright:rmicse}) as a metaphor for the visualization of evolution of an optimization process.
Usually, on the basis of a $n$-dimensional search space, an extra dimension is added which represents the fitness of each solution. So, this $(n+1)-$dimension space can be interpreted like a landscape with valleys and peaks. This landscape is more or less rough according to the complexity of the problem. However, this view of the search space presents some limitations. It is hard to visualize a search space of dimension higher than 2; the concept of neighboring, induced by a distance, an operator or an heuristic, is not easily perceptible; it is difficult to locate, to count or to characterize the set of local optima, as soon as the landscape becomes rough; barriers of fitness between basins of attraction are not always highlighted and dynamics of search heuristics cannot be directly tracked on the landscape.

%%% ------------------------------------------------------------------
\section{The fitness cloud}
This section presents a complementary "point of view" to the geographical metaphor of landscape. The search space is noted $S$ and the fitness function $f$ is defined on $S$.

\subsection{Bordering fitness}

Two solutions are regarded as neighbor if there is a transformation related to search heuristics or such an operator, which allows "to pass" from one solution to the other one.
Let $s$ be a solution in the search space, its \textit{bordering fitness} $\tilde f(s)$, is defined as the fitness of a particular neighbor of $s$. The choice of one neighbor depends on the search heuristic only and we assume this choice to be unique. 

\subsection{Definition}

For each solution in the solution space, a single point is plotted; the abscissa is its fitness and the ordinate is its bordering fitness. Thus, we obtain a scatterplot which informs about the correlation between fitness and bordering fitness (the so-called \textit{Fitness Cloud} or FC).
Formally, $FC=\{ (f(x),\tilde{f}(x))\ |\ x \in S \}$.
A \textit{set of neutrality} of fitness $\varphi$, so-called $S_{\varphi}$, is the set of solutions that have the fitness $\varphi$. Such a set corresponds to one abscissa in the fitness/fitness plan; according to this abscissa, a vertical slice from the cloud represents all the fitness values one can reach from this set of neutrality. From a given bordering fitness value $\tilde f$, an horizontal slice represents all the fitness values from which one can reach $\tilde f$. 

To visualize the shape of the fitness cloud, we plot the three subsets of $FC$: 
$FC_{min} = \{ {\left( {\varphi ,\tilde \varphi } \right)\ |\ \varphi  \in f\left( {S} \right),\ \tilde \varphi  = \mathop {\min }\limits_{x \in S_\varphi  } \tilde f\left( x \right)} \}$, 
$FC_{max} = \{ {({\varphi ,\tilde \varphi })\ |\ \varphi  \in f( {S}),\ \tilde \varphi  = \mathop {\max }\limits_{x \in S_\varphi  } \tilde f(x)}\}$ and
$FC_{mean} = \{ {\left( {\varphi ,\tilde \varphi } \right)\ |\  \varphi  \in f\left( {S} \right),\ \tilde \varphi  = \mathop {mean}\limits_{x \in S_\varphi } \tilde f\left( x \right)} \}$.

\subsection{Evolvability on fitness cloud}

\textit{Evolvability} is defined by \cite{WA-AL} as "the ability of random variations to sometimes produce improvement".
There are three specific fitness values\footnote{Existence of which depends on both the problem and the heuristic} (respectively $\alpha, \beta, \gamma$) corresponding to the intersection of the curves (respectively $FC_{min}$, $FC_{mean}$ and $FC_{max}$) with the diagonal line ($\tilde{f}=f$). So, according to the fitness level $\varphi$, four cases can be enumerated (see fig.~\ref{NK_25_20_best1_run}):
\begin{enumerate} 

        \item $\varphi \le \alpha$: bordering fitness is always higher than fitness; applying the heuristic confers selective advantage.
        \item $\alpha  < \varphi \le \beta$: the mean bordering fitness is higher than fitness. Thus, on average the heuristic is selectively advantageous.

        \item $\beta  < \varphi \le \gamma$: the mean bordering fitness is lower than fitness. Thus, on average the heuristic is selectively deleterious.
        \item $\gamma  < \varphi$: bordering fitness is always lower than fitness. The heuristic is selectively deleterious.
\end{enumerate}

\section{Experimental results on NK-landscape}

The search space is the set of bit-string of length $N=25$. Two strings are neighbors if their Hamming distance is one. All experiments are led on the same instance of NK-landscape with $K=20$. Datas are collected from an exhaustive enumeration of the search space\footnote{A sampling of the search also could be realize if it is large}. Practically two fitness values are taken as equal if they both stand in the same interval of size $0.002$. 

\subsection{Whole Fitness Cloud}

\begin{figure}[ht]
        \begin{center}  
    	\includegraphics[width=0.33\textwidth]{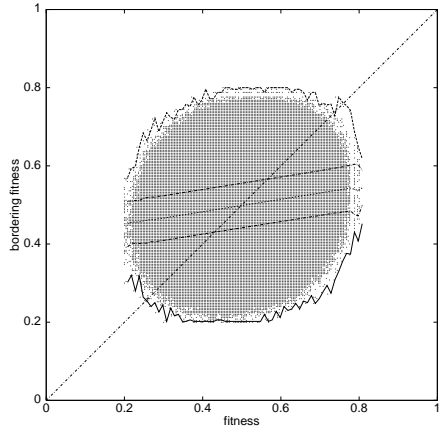}
        \end{center}
        \caption{The whole fitness cloud of NK-landscape with $N=25$ and $K=20$: the fitness cloud ($FC$) and it shape ($FC_{min}$, $FC_{max}$ and $FC_{mean}$ with standart-deviation) under the hamming neighborhood. The $FC_{mean}$ curve is roughly a line.}
        \label{NK_25_20_nuage_contourHamming}
\end{figure}

We draw scatterplot, the so-called whole fitness cloud including, for each string of the search space, all the points in the hamming neighborhood (see fig.\ref{NK_25_20_nuage_contourHamming}).
As the density of points on the scatterplot gives little information on dispersion, a standard deviation is plotted on both side of the $mean$ curve. 

The fact that the $FC_{mean}$ curve computing on the whole scatterplot is roughly a line (see fig.\ref{NK_25_20_nuage_contourHamming}) confirms the results from Weinberger \cite{WEI:90}:
$\tilde f_{mean}  = \left( {1 - \frac{{K + 1}}{N}} \right)f + \left( {\frac{{K + 1}}{N}} \right)0.5$
As reported by \cite{SMI:01}, let us note that the slope coefficient ${1-\frac{{K + 1}}{N}}$ is the offspring-parent fitnesses correlation. 

\subsection{Hill climbing}

A greedy hill climbing heuristic (so-called GHC) is used. 

\begin{figure}[ht]
        \begin{center}  
        \includegraphics[width=0.33\textwidth]{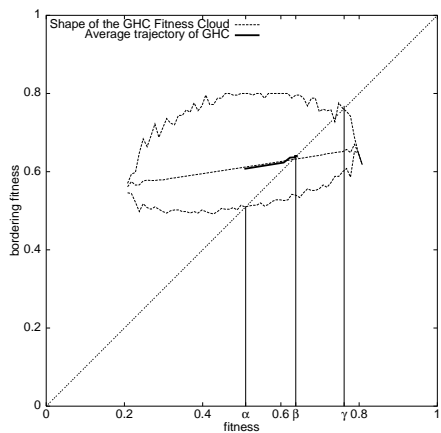}
        \end{center}
        \caption{The thin line is the shape of fitness cloud ($FC_{min}$, $FC_{max}$ and $FC_{mean}$ with standart-deviation) under GHC of NK-landscape with $N=25$ and $K=20$. The line is the average trajectory of GHC}
        \label{NK_25_20_best1_run}
\end{figure}

\subsubsection{FC, local optima and epistasis} 

A local optimum is a point in the landscape which is higher than any of the points which immediately surround it.
For such a point, the best possible fitness over its neighbourhood is less fit than it; so, its bordering fitness is lower than its fitness. Within the cloud, local optima fit points under the diagonal line (see fig.~\ref{NK_25_20_best1_run}). Such a localization gives insight on the amount and the fitnesses of local optima.

Examining the fitness cloud, the set $FC_{mean}$ seems to be coarsely supported by a line (see fig.~\ref{NK_25_20_best1_run}). As for the whole fitness cloud, we can prove that $FC_{mean}$ is a line with the same slope of $1 - \frac{K+1}{N}$ and the Y-intercept is a constant which depends on $N$ and $K$.

\subsubsection{Dynamics on the Fitness Cloud under GHC}

We conjecture that the $\beta$ fitness level is a \textit{barrier of fitness}. This means that, applying GHC heuristic from a random initial solution, on average the search process breaks off around $\beta$. To validate this hypothesis we conduct a number of experiments on the NK-landscape with GHC: the search heuristic is run over 100 generations to collect information on the dynamics as the list of successive points $(f,\tilde f)$ encountered during the search process. All the experimental datas collected from 70 such runs allows to build an \textit{average trajectory}. As expected this trajectory starts on the ${FC_{mean}}$ line with a fitness near to $0.5$\footnote{Fitness of a random initial solution is closed to the mean fitness over the search space ($\bar f=0.5$)}, and then roughly follows the ${FC_{mean}}$ line to finally breaks off around the $({\beta;\beta})$ point (see fig.~\ref{NK_25_20_best1_run}). Therefore, examining the fitness cloud allows to predict the average long-term behavior for GHC at fitness level.

\section*{Conclusion}

In this paper we have presented the \textit{Fitness Cloud} as a complementary viewpoint to the \textit{Fitness Landscape} metaphor. FC is a 2-d representation where the topology induced by an heuristic is directly taken into account. Our analytical and empirical results suggest that FC allows us to characterize the set of local optima and barriers of fitness too. In others experiments on Simulated Annealing, we have established that FC can predict the barriers of fitness.
In such a context, we believe the FC can be used beneficially to track the dynamic and to predict the average behavior of the search process. To change the metaphor from landscape to cloud leads change to the picture from that of a point getting stuck on a local peak to that of a point pulled towards a particular set of neutrality.\\

\bibliography{Biblio}

\end{document}